\documentclass[runningheads]{llncs}
\usepackage{times}
\usepackage{latexsym}
\usepackage{graphicx}
\usepackage{float}
\usepackage{color, colortbl}
\usepackage{xcolor}
\definecolor{Gray}{gray}{0.9}
\usepackage{makecell}
\usepackage{lipsum}
\usepackage{float}
\usepackage{placeins}
\usepackage{hyperref}
\usepackage{epstopdf}
\usepackage{algorithm}
\usepackage{amsmath}
\usepackage{amssymb}
\usepackage{pifont}
\usepackage{multirow}
\usepackage{subcaption}
\begin{document}
\title{"Explain Thyself Bully": Sentiment Aided Cyberbullying Detection with Explanation}
\titlerunning{Explain Thyself Bully}
\author{Krishanu Maity\inst{1\star}\and
Prince Jha\inst{1}\thanks{Both authors contributed equally to this research.}\and
Raghav Jain\inst{1}\and
Sriparna Saha\inst{1} \and 
Pushpak	Bhattacharyya\inst{2}}
\authorrunning{Maity et al.}
\institute{Department of Computer Science and Engineering, Indian Institute of Technology Patna, India \and
Department of Computer Science and Engineering, Indian Institute of Technology Bombay, India\\
\email{\{krishanu\_2021cs19,princekumar\_1901cs42\}@iitp.ac.in}\\
\email{\{raghavjain106,sriparna.saha,pushpakbh\}@gmail.com}\\}
\maketitle              % typeset the header of the contribution
\begin{abstract}
Cyberbullying has become a big issue with the popularity of different social media networks and online communication apps.  While plenty of research is going on to develop better models for cyberbullying detection in monolingual language, there is very little research on the code-mixed languages and explainability aspect of cyberbullying. Recent laws like "right to explanations" of General Data Protection Regulation, have spurred research in developing interpretable models rather than focusing on performance. Motivated by this we develop the first interpretable  multi-task model called {\em mExCB} for automatic cyberbullying detection from code-mixed languages which can simultaneously solve several tasks, cyberbullying detection, explanation/rationale identification, target group detection and sentiment analysis. We have introduced {\em BullyExplain}, the first benchmark dataset for explainable cyberbullying detection in code-mixed language. Each post in {\em BullyExplain} dataset is annotated with four labels, i.e., {\em bully label, sentiment label, target and rationales (explainability)}, i.e., which phrases are being responsible for annotating the post as a bully. The proposed multitask framework (mExCB) based on CNN and GRU with word and sub-sentence (SS) level attention is able to outperform several baselines and state of the art
models when applied on {\em BullyExplain} dataset.\footnote{The code and dataset are available at \url{https://github.com/Jhaprince/BullyExplain}.} \\
{\bf Disclaimer:} The article contains offensive text and
profanity. This is owing to the nature of the work, and do not reflect any opinion or stand of the authors.
%Cyberbullying
%Sentiment
%Emotion
%Code-Mixed(Hindi+English)
%Multi-task

\keywords{Cyberbullying  \and Sentiment \and Code-Mixed \and Multi-task \and Explainability.}
\end{abstract}
\section{Introduction} 
Cyberbullying~\cite{cyb1} is described as an aggressive, deliberate act committed against people using computers, mobile phones, and other electronic gadgets by one individual or a group of individuals. According to the Pew Research Center, 40\% of social media users have experienced some sort of cyberbullying\footnote{\url{https://www.pewresearch.org/internet/2017/07/11/online-harassment-2017/}}. Victims of cyberbullying may endure sadness, anxiety, low self-esteem,  transient fear and suicidal thinking~\cite{sticca2013longitudinal}. According to the report of the National Records Crime Bureau, \cite{ncrb} “A total of 50,035 cases were registered under Cyber Crimes, showing an increase of 11.8\% in registration over 2019
(44,735 cases)”. The objective of minimizing these harmful consequences emphasizes the necessity of developing techniques for detecting, interpreting, and preventing cyberbullying. 

Over the last decade, most of the research on cyberbullying detection has been conducted on monolingual social media data using traditional machine learning~\cite{dadvar2014experts,rwml1,rwml2} and deep learning models~\cite{agrawal2018deep,waseem2016hateful,rwml4}. The contingency of code-mixing is increasing very rapidly. Over 50M tweets were analyzed by~\cite{rijhwani2017estimating}, in which 3.5\% of tweets are code-mixed. Hence this should be our primary focus at this point in time. Code-mixing is a linguistic marvel in which two or more languages are employed in speech alternately~\cite{cm1}. Recently, people have started working on offensive post-detection in code-mixed languages like aggression detection~\cite{kumar2018aggression}, hate-speech detection \cite{rwcm3}, and cyberbullying detection~\cite{maity2021multi,maity2022mtbullygnn}. However, those researchers mostly concentrated on improving the performance of cyberbullying detection using various models, without giving any insight or analysis into the explainability. With the introduction of explainable artificial intelligence (AI) \cite{gunning2019xai}, it is now a requirement to provide explanations/interpretations for any decision taken by a  machine learning algorithm. This helps in building trust and confidence when putting AI models into production.
Moreover, in Europe, legislation such as the General Data Protection Regulation (GDPR)~\cite{regulation2016regulation} has just introduced a "right to explanation" law. Hence there is an urge in developing  interpretable models rather than only focusing on improving performance by increasing the complexity of the models.

In this paper, we have developed an explainable cyberbullying dataset and a hierarchical attention-based multitask model to solve four tasks simultaneously, i.e., Cyberbullying Detection (CD), Sentiment Analysis (SA), Target Identification (TI), and Detection of Rationales (RD). To develop an explainable code-mixed cyberbullying dataset, we have re-annotated the existing {\em BullySent} dataset~\cite{maity2021multi} with the target class ( Religion, Sexual-Orientation, Attacking-Relatives-and-Friends, Organization, Community, Profession and Miscellaneous) and highlighted parts of the text that could justify the classification decision. If the post is non-bully, the rationales are not marked and the target class is selected as NA (Not Applicable). This study focuses on applying rationales, which are fragments of text from a source text that justify a classification decision. Commonsense explanations~\cite{rajani2019explain}, e-SNLI~\cite{camburu2018snli}, and various other tasks~\cite{deyoung2019eraser} have all employed such rationales. If these rationales are valid explanations for decisions, models that are trained to follow them might become more human-like in their decision-making and thus are more trustworthy, transparent, and reliable. 

A person's sentiment heavily influences the intended content~\cite{lewis2010handbook}. In a multitask (MT) paradigm, the task of sentiment analysis (SA) has often been considered as an auxiliary task to increase the performance of primary tasks (such as cyberbullying detection (CD)~\cite{maity2021multi}, Complaint Identification~\cite{singh2021multitask} and tweet act classification (TAC)~\cite{saha2021multitask}). In the current set-up,  sentiment analysis and rationale identification are treated as secondary tasks as the presence of rationale with sentiment information of a post certainly helps identify bully samples more accurately. Our proposed multitask model incorporates a Bi-directional Gated Recurrent  Unit (Bi-GRU), Convolutional Neural Network (CNN), and Self-attention in both word level and sub-sentence (SS) level representation of input sentences. CNNs perform well for extracting local and position-invariant features~\cite{yin2017comparative}. In contrast, RNNs are better for long-range semantic dependency-based tasks (machine translation, language modeling) than some local key phrases. The intuition behind the usage of CNN and Bi-GRUs in both word level and sub-sentence level is to efficiently handle code-mixed data, which is noisier (spelling variation, abbreviation, no specific grammatical rules) than monolingual data.

The following are the major contributions of the current paper:
\begin{itemize}
  \item We investigate two new issues: (i) explainable cyberbullying detection in code-mixed text and (ii) detection of their targets. 
  
  \item {\em BullyExplain,} a new benchmark dataset for explainable cyberbullying detection with target identification in the code-mixed scenario, has been developed.
  
  \item To simultaneously solve four tasks (Bully, Sentiment, Rationales, Target), an end-to-end deep multitask framework (mExCB) based on word and subsentence label attention has been proposed. 
  
  \item Experimental results illustrate that the usages of rationales and sentiment information significantly enhance the performance of the main task, i.e., cyberbullying detection.
 
\end{itemize}

\section{Related Works}
Text mining and NLP paradigms have been used to investigate numerous subjects linked to cyberbullying detection, including identifying online sexual predators, vandalism detection, and detection of internet abuse and cyberterrorism. Although the associated research described below inspires cyberbullying detection, their methods do not consider the explainability part, which is very much needed for any AI/ML task.

\subsection{Works on Monolingual Data} Dinakar et al.~\cite{rwml1} investigated cyberbullying detection using a corpus of 4500 YouTube comments and various binary and multiclass classifiers. The SVM classifier attained an overall accuracy of 66.70\%, while the Naive Bayes classifier attained an accuracy of 63\%. Authors in~\cite{rwml2} developed a Cyberbullying dataset by collecting data from Formspring.me and finally achieved 78.5\% accuracy by applying C4.5 decision tree algorithm using  Weka tool kit. In 2020, Balakrishnan et al.~\cite{rwml5} developed a model based on different machine learning approaches and psychological features of twitter users for cyberbullying detection. They found that combining personality and sentiment characteristics with baseline features (text, user, and network) enhances cyberbullying detection accuracy. CyberBERT, a BERT based framework developed by~\cite{paul2020cyberbert} achieved the state-of-the-art results on Formspring (12k posts), Twitter (16k posts), and Wikipedia (100k posts) dataset.

\subsection{Works on Code-mixed Data} In \cite{maity2021bert}, the authors attained 79.28\% accuracy in cyberbullying detection by applying CNN, BERT, GRU, and Capsule Networks on their introduced code-mixed Indian language dataset. For identifying hate speech from Hindi-English code mixed data, the deep learning-based domain-specific word embedding model in~\cite{rwcm5}  outperforms the base model by 12\% in terms of F1 score. Authors in~\cite{rwcm2} created an aggression-annotated corpus that included 21k Facebook comments and 18k tweets written in Hindi-English code-mixed language.

\subsection{Works on Rationales:} 
Zaidan et al.~\cite{zaidan2007using} proposed the concept of rationales, in which human annotators  underlined a section of text that supported their tagging decision. Authors have examined that the usages of these rationales certainly improved sentiment classification performance. Mathew et al.~\cite{mathew2020hatexplain} introduced HateXplain, a benchmark dataset for hate speech detection. They found that models that are trained using human rationales performed better at decreasing inadvertent bias against target communities.  Karim et al.~\cite{karim2021deephateexplainer} developed an explainable hate speech detection approach (DeepHateExplainer) in Bengali based on different variants of transformer architectures (BERT-base, mMERT, XLM-RoBERTa).  They have provided explainability by highlighting the most important words for which the sentence is labeled as hate speech.

\subsection{Works on Sentiment Aware Multitasking}
There are some works in the literature where sentiment analysis is treated as an auxiliary task to boost the performance of the main task. Saha et al.~\cite{saha2021multitask} proposed a multi-modal tweet act classification (TAC) framework based on an ensemble adversarial learning strategy where sentiment analysis acts as a secondary task.~\cite{singh2021multitask} developed a multi-task model based on affective space as a commonsense knowledge. They achieved a good accuracy of 83.73\% and 69.01\% for the complaint identification and sentiment analysis tasks, respectively. Maity et al.~\cite{maity2021multi} created a Hindi-English code-mixed dataset for cyberbullying detection. Based on BERT and VecMap embeddings, they developed an attention-based deep multitask framework and achieved state-of-the-art performance for sentiment and cyberbullying detection tasks. 

After performing an in-depth literature review, it can be concluded that there is no work on explainable cyberbullying detection in the code-mixed setting. 
In this paper, we attempt to fill this research gap. 

\section{{\em BullyExplain} Dataset Development}
This section details the data set created for developing the explainable cyberbullying detection technique in a code-mixed setting.

\subsection{Data Collection}
To begin, we reviewed the literature for the existing code-mixed cyberbullying datasets. We found two cyberbullying datasets~\cite{maity2021bert},~\cite{maity2021multi} in Hindi-English code-mixed tweets. We selected the BullySent~\cite{maity2021multi} dataset for further annotation with rationales and target class as it was previously annotated with the bully and sentiment labels. 

\subsection{Annotation training}
The annotation was led by three Ph.D. scholars with adequate knowledge and expertise in cyberbullying, hate speech, and offensive content and performed by three undergraduate students with proficiency in both Hindi and English. First, a group of undergraduate computer science students were voluntarily hired through the department email list and compensated through gift vouchers and honorarium. Previously each post in BullySent~\cite{maity2021multi} dataset was annotated with bully class (Bully / non-bully) and sentiment class (Positive / Neutral / Negative). For rationales and target annotation, we have considered only the bully tweets. For annotation training, we required gold standard samples annotated with rationale and Target labels. Our expert annotators randomly selected 300 memes and highlighted the words (rationales) for the textual explanation and tag a suitable target class. For rational annotation, we have followed the same strategy as mentioned in~\cite{mathew2020hatexplain}. Each word in a tweet was marked with either 0 or 1, where 1 means rationale. We have considered seven target classes (Religion, Sexual-Orientation, Attacking-Relatives-and-Friends, Organization, Community, Profession and Miscellaneous) as mentioned in~\cite{mathew2020hatexplain} and \cite{pramanick2021detecting}. Later expert annotators discussed each other and resolved the differences to create 300 gold standard samples with rationale and target annotations. We divide these 300 annotated examples into three sets, 100 rationale annotations each,  to carry out three-phase training. After the completion of every phase, expert annotators met with novice annotators to correct the wrong annotations, and simultaneously annotation guidelines were also renewed. After completing the third round of training, the top three annotators were selected to annotate the entire dataset. 

\subsection{Main annotation}
We used the open-source platform Docanno\footnote{\url{https://github.com/doccano/doccano}} deployed on a Heroku instance for main annotation where each qualified annotator was provided with a secure account to annotate and track their progress exclusively. We initiated our main annotation process with a small batch of 100 memes and later raised it to 500 memes as the annotators became well-experienced with the tasks. We tried to maintain the annotators’ agreement by correcting some errors they made in the previous batch. On completion of each set of annotations, final rationale labels were decided by the majority voting method. If the selections of three annotators vary, we enlist the help of an expert annotator to break the tie. We also directed annotators to annotate the posts without regard for any particular demography, religion, or other factors. We use the Fleiss' Kappa score~\cite{fleiss1971measuring} to calculate the inter-annotator agreement (IAA) to affirm the annotation quality. IAA obtained scores of 0.74 and 0.71 for the rationales detection (RD) and Target Identification (TI) tasks, respectively, signifying the dataset being of acceptable quality. Some samples from the {\em BullyExplain} dataset are shown in Table~\ref{tab:my-samples}.

\subsubsection{Ethics note}
Repetitive consumption of online abuse could distress mental health conditions \cite{ybarra2006examining}. Therefore, we advised annotators to take periodic breaks and not do the annotations in one sitting. Besides, we had weekly meetings with them to ensure the annotations did not have any adverse effect on their mental health.

\begin{table*}[hbt]
\centering
\caption{Some samples from annotated {\em BullyExplain} dataset; The green highlights mark rationales tokens. AFR  - "Attacking-Relatives-and-Friends"}
\label{tab:my-samples}
\scalebox{0.72}{
\begin{tabular}{llll}
\hline
\textbf{Tweet} &
  \textbf{\begin{tabular}[c]{@{}l@{}}Bully\\ Label\end{tabular}} &
  \textbf{\begin{tabular}[c]{@{}l@{}}Sentiment\\ Label\end{tabular}} &
  \textbf{Target} \\ \hline
\begin{tabular}[c]{@{}l@{}}T1: \colorbox{green}{Larkyaaan} toh jaisyyy bht hi \colorbox{green}{phalwaan} hoti. Ak \colorbox{green}{chipkali} ko dekh kr\\ tm logon ka \colorbox{green}{sans rukk jataa}\\ Translation: Yes, I know the \colorbox{green}{girls} are \colorbox{green}{brave}. Even a tiny \colorbox{green}{lizard} can \colorbox{green}{stop their heart}\end{tabular} &
  Bully &
  Negative &
  Sexual-Orientation \\
\begin{tabular}[c]{@{}l@{}}T2: My friend called me \colorbox{green}{moti} and I instantly replied with "tere baap ka khati hoon"\\ Translation: My friend called me \colorbox{green}{fatso} and my instant reply was \\ "does your father bear my cost?"\end{tabular} &
  Bully &
  Negative &
  AFR \\
\begin{tabular}[c]{@{}l@{}}T3: Laal phool gulaab phool shahrukh bhaiya beautiful \\ Translation: Red flowers are roses, brother Shahrukh is beautiful.

\end{tabular} &
  Non-bully &
  Positive &
  NA\\ \hline
\end{tabular}}
\end{table*}

\begin{figure}[hbt]
	\centering
	\includegraphics[height = 6 cm, width = 10 cm]{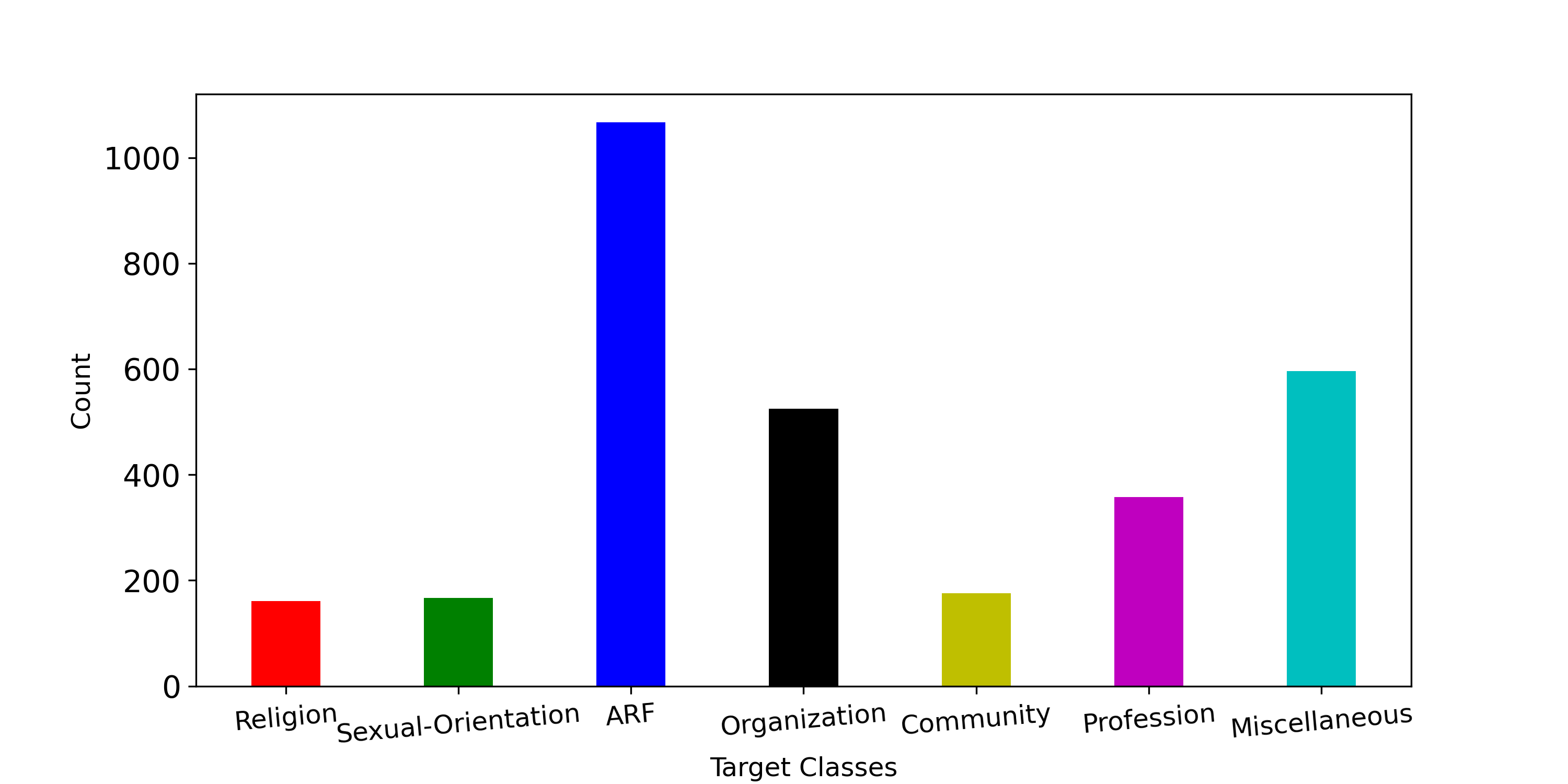}
	\caption {Statistics of target class in our developed dataset.}
	\label{fig:target}
	
\end{figure}

\subsection{Dataset Statistics}
In the {\em BullyExplain} dataset, the average number of highlighted words per post is 4.97, with an average of 23.15 words per tweet. {\em maa} (mother), {\em randi} (whore), and {\em gandu} (asshole) are the top three content words for bully highlights, appearing in 33.23 percent of all bully posts. The total number of samples in the {\em BullyExplain} dataset is 6084, where 3034 samples belong to the non-bully class and the remaining 3050 samples are marked as bully. The number of tweets with positive and neutral sentiments are 1,536 and 1,327, respectively, while the remaining tweets have negative sentiments. Figure~\ref{fig:target} shows the statistics of the Target class in  {\em BullyExplain} dataset.  From figure~\ref{fig:target}, we can observe that approximately one-third of total bully samples (3050) belong to the Attacking-Relatives-and-Friends (ARF) category (1067).  This statistic reveals the nature of cyberbullying problem, where the victim's relatives and friends are the targets most of the time.

\section{Explainable Cyberbullying Detection}
\begin{figure*}[t]
	\centering
	\includegraphics[height = 9 cm, width = 12 cm]{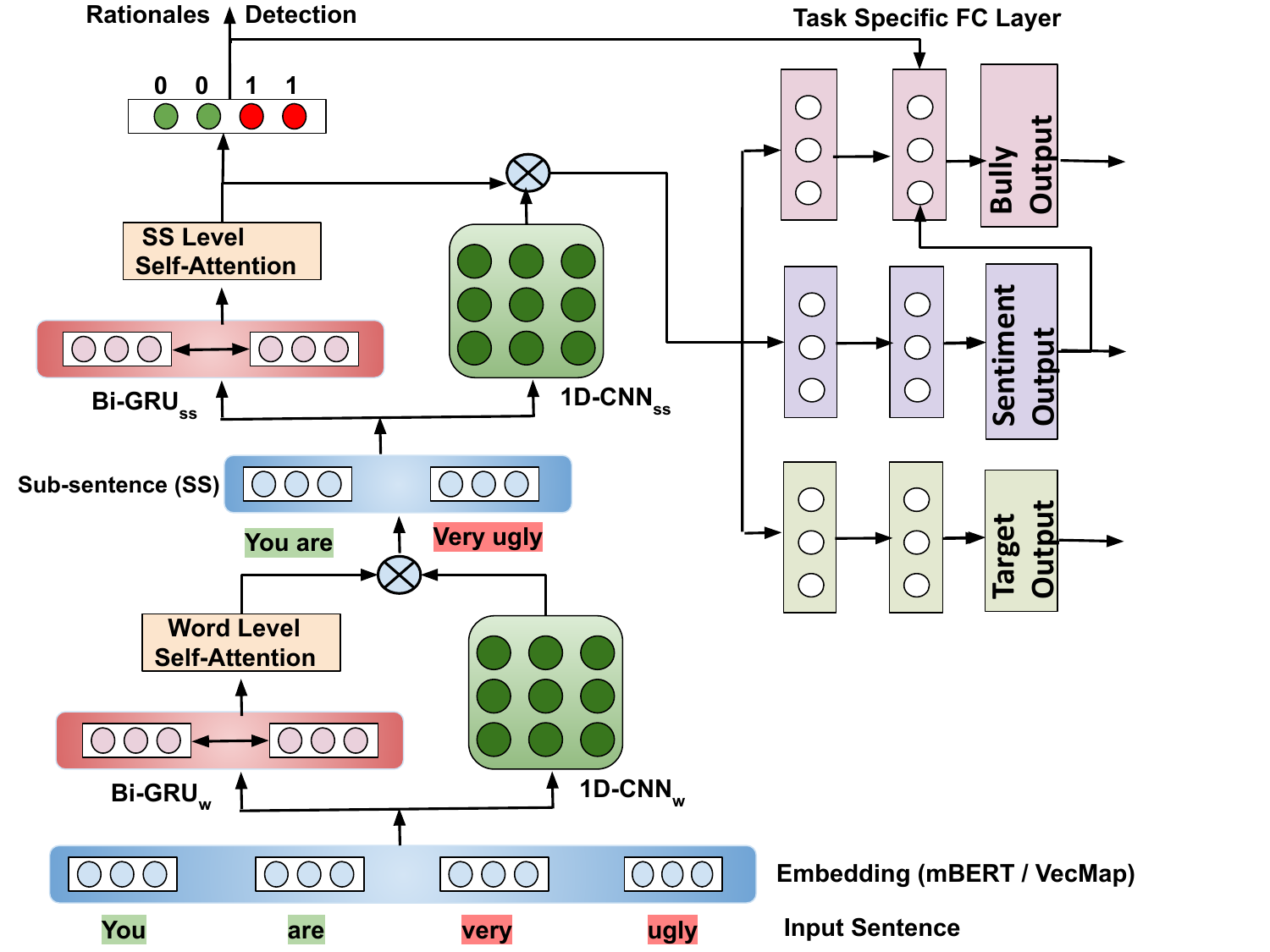}
	\caption {Proposed Multitask Framework for Explainable
Cyberbullying Detection, {\em mExCB}, architecture.}
	\label{fig:archi}
	
\end{figure*}

This section presents our proposed "mExCB" model, shown in Figure~\ref{fig:archi}, for explainable cyberbullying detection.  We utilized CNN and bi-GRU in both word and sub-sentence levels, along with self-attention, to make a robust end-to-end multitask deep learning model. 
\subsection{Text Embedding Generation} 
To generate the embedding of input sentence $S$ (say) containing $N$ number of tokens, we have experimented with BERT and VecMap.

(i) {\bf BERT}~\cite{devlin2018bert} is a language model based on bidirectional transformer encoder with a multi-head self-attention mechanism. The sentences in our dataset are written in Hindi-English code-mixed form, so we choose mBERT (Multilingual BERT) pre-trained in 104 different languages, including Hindi and English. We have considered the sequence output from BERT, where each word of the input sentence has a 768-dimensional vector representation. 

(ii) {\bf VecMap}~\cite{artetxe2017learning} is a multilingual word embedding mapping method. The main idea behind VecMap is to consider pretrained source and destination embeddings separately as inputs and align them in a shared vector space where related words are clustered together using a linear transformation matrix. As the inputs of VecMap, we have considered Fasttext~\cite{grave2018learning} Hindi and English monolingual embeddings because FastText employs the character level in represented words into the vectors, unlike word2vec and Glove, which use word-level representations.

\subsection{Feature Extraction}
Bi-GRU and CNN have been employed to extract the hidden features from the input word embedding. 
(i) {\bf Bi-GRU}~\cite{cho2014properties} learns long term context-dependent semantic features into hidden states by sequentially encoding the embedding vectors, $e$, as 
\begin{equation}
\overrightarrow{h}_{t} = \overrightarrow{GRU}_{fd}(e_{t}, \;h_{t-1}),\overleftarrow{h}_{t} = \overleftarrow{GRU}_{bd}(e_{t}, \;h_{t+1})
\end{equation}
Where $\overrightarrow{h}_{t}$ and $\overleftarrow{h}_{t}$ are the forward and backward hidden states, respectively.
The final hidden state representation for the input sentence is obtained as,
%$\begin{equation}
$H_{e} = \left[ h_{1}, h_{2},h_{3},....h_{N} \right]$\\
%\end{equation}
where ${h}_{t} = \overrightarrow{h}_{t}, \overleftarrow{h}_{t}$ and 
$H_{e} \in \mathbb{R}^{N \times 2D_{h} }$. The number of hidden units in GRU is $D_{h}$.

(ii) {\bf CNN}~\cite{kim2014convolutional} effectively captures abstract representations that reflect semantic meaning at various positions in a text. To obtain N-grams feature map, $\bf c \in \mathbb{R}^{n - k_{1} + 1}$ using  filter $F \in \mathbb{R}^{k_{1} \times d}$, we perform convolution operation, an element-wise dot product over each possible word-window, $W_{j:j+k_{1}-1}$. Each element $c_{j}$ of feature map $\bf c$ is generated after convolution by 
\begin{equation}
\label{conv}
c_{j} = f(w_{j:j+k_{1}-1} * F_{a} + b)
\end{equation} 
Where $f$ is a non linear activation function and $b$ is the bias. Then we perform max pooling operation on $\bf c$. After applying $F$ distinct filters of the same N-gram size, $F$ feature maps will be generated, which can then be rearranged as 
%\begin{equation}
$${\bf C = [c_{1}, c_{2}, c_{3}, ......c_{F}] }$$
%\end{equation}

{\bf Self-attention }~\cite{vaswani2017attention} has been employed to determine the impact of other words on the current word by mapping a query and a set of key-value pairs to an output. Outputs from Bi-GRU are passed through three fully connected (FC) layers, namely queries(Q), keys(K), and values(V) of dimension $ D_{sa}$.  Self-attention scores (SAT) are computed as follows:
\begin{equation}
SAT_{i} = softmax(Q_{i}K_{i}^{T})V_{i}
\label{eq3}
\end{equation}

where $ SAT_{i} \in \mathbb{R}^{N \times D_{sa} }$. 

\subsection{mExCB :Multitask Framework for Explainable
Cyberbullying Detection} We delineate the end-to-end process of the mExCB model as follows:
\begin{enumerate}
\item We initially compute the embedding for each word in a sentence, $ S =\left \{ s_{1}, s_{2} , .....s_{N}\right \}$.
$ E_{N \times D_{e}}^{w} = Embedding_{BERT/VecMap}(S) $ where $D_{e}$ is the embedding dimension (for BERT $D_{e}$ = 768 and for VecMap $D_{e}$ = 300).
\item Word Level Representation: $E_{N \times D_{e}}^{w}$ has been passed through both Bi-GRU followed by Self-Attention and CNN layers to get word-level hidden features.\\
$ H_{N \times 2D_{h}}^{w} = GRU(E) $; where $D_{h}$ = 128.\\
$ A_{N \times D_{sa}}^{w} = SAT(H) $; where $D_{sa}$ = 200.\\
Next, we take the mean of self-attenuated $N$ vectors generated by GRU+SAT at the interval of $l$, where $l$ is the number of words to generate a sub-sentence (SS). \\
$ G_{P \times D_{sa}}^{w} = AVG_{l}(A) $; where $P$ = $N/l$.\\
$ C_{P \times F}^{w} = CNN(E) $; where $F$ = 200 

\item Sub-sentence Level Representation: We have added $ G^{w}$ and $ C^{w}$ to get the sub-sentence level embedding, $ E_{P \times D_{sa}}^{ss}$. We followed similar steps as mentioned during the word-level representation to generate the sub-sentence level convoluted features, $ C_{1 \times F}^{ss}$, and attenuated recurrent (GRU+SAT) features, $ G_{1 \times D_{sa}}^{ss}$. 

\item $G^{ss}$ and $C^{ss}$ are concatenated to get the final representation, $E^{s}$, of the given input, $S$. Up to this, all the layers are common for all the tasks, which basically helps in sharing task-specific information.
\item Task Specific Layers: We have three task-specific fully connected (FC) layers ($FC_{1}-FC_{2}$-Softmax) followed by the corresponding output layer to simultaneously solve three tasks (Bully, Sentiment, and Target). For Rationales identification, we have fed $G^{ss}$ into a sigmoid output layer which returns a binary encoding vector as an output. Each FC layer has 100 neurons.  Further, outputs generated by the rationales identification and sentiment channels have been added to the last FC layer ($FC_{2}$) of the bully channel to examine how sentiment and rationales information helps in boosting the performance of cyberbullying detection.

\end{enumerate}

\subsection{Loss Function} Categorical cross-entropy~\cite{zhang2018generalized} has been employed as an individual loss function for all the tasks. The final loss function, $Loss_{f}$ is dependent on $M$ task-specific individual losses,  $Loss_{s}^k$, as follows: 
\begin{equation}
Loss_{f} =  \sum_{k=1}^{M}\beta _{k}Loss_{s}^{k}
\end{equation}
 The variable $\beta$ is a hyperparameter which ranges from 0 to 1, defines the loss weights that characterise the per task loss-share to the total loss.

\section{Experimental Results and Analysis}
This section describes the outcomes of various baseline models and our proposed model, tested on the {\em BullyExplain} dataset. The experiments are intended to address the following research questions:
\begin{itemize}
\item \textbf{RQ1}: How does multi-tasking help in enhancing  the performance of CD task?

\item \textbf{RQ2}: What is the effect of different task combinations in our framework?

\item \textbf{RQ3}: What is the motivation for keeping both CNN and GRU in our proposed framework?

\item \textbf{RQ4}: Why do we use only the embeddings from pre-trained mBERT, but do not fine-tune the model itself for the multi-task set-up?

\item \textbf{RQ5}: To handle noisy code-mixed data, which embedding is better, BERT or VecMap?

\end{itemize}
We performed stratified 10-fold cross-validation on our dataset and reported the mean metrics scores as done in \cite{maity2021multi}. During validation, we experimented with different network configurations and obtained optimal performance with batch size = 32, activation function=ReLu, dropout rate= 0.25, learning rate= 1e-4, epoch= 20. We used Adam optimizer with a weight\_decay=1e-3 ( for avoiding overfitting) for training. We set the value of $\beta$ for the bully, rationales, sentiment, and target tasks for all the multi-task experiments by 1, 0.75, 0.66, and 0.50, respectively. All our experiments are performed on a hybrid cluster of multiple GPUs comprised of RTX 2080Ti.

\subsection{Baselines Setup}
We have experimented with different standard baseline techniques like CNN-GRU, BiRNN, BiRNN-Attention, and BERT-finetune, as mentioned in \cite{mathew2020hatexplain}.  To investigate why both CNN and GRU are important in our model, we have performed an ablation study by adding other baselines (5 and 6) as follows:

\begin{enumerate}
\item \textbf{CNN-GRU}:
The input is sent through a 1D CNN with window sizes of 2, 3, and 4, each with 100 filters. We employ the GRU layer for the RNN portion and then max-pool the output representation from the GRU architecture's hidden layers. This hidden layer is processed via a fully connected layer to output the prediction logits. 

\item \textbf{BiRNN}: We pass the input features to BiRNN. The obtained hidden representation from BiRNN is sent to fully connected layers to obtain output. 

\item \textbf{BiRNN-Attention}: This model differs from BiRNN model only in terms of the attention layer.

\item \textbf{BERT-finetune}: BERT's pooled output with dimension 768 was fed to a softmax output layer.

\item {\bf mExCB$_{CNN}$}:  In this architecture, we keep two 1D CNNs for word level and sub-sentence level encoding one after the other. The contextual hidden representation obtained from CNN is passed through multitasking channels.

\item {\bf mExCB$_{GRU}$}: In this baseline, input features are passed through two Bi-GRU+SAT layers. Outputs from sub-sentence level representation are passed through multitasking channels.

\end{enumerate}

There are four multitask variants based on how many tasks we want to solve simultaneously. As we have four tasks and our main objective is explainable cyberbullying detection, CD and RD are kept common for any multitask variants. So we have four multitask variants, i.e., CD+RD, CD+RD+SA, CD+Rd+TI, and CD+RD+SA+TI. For the CD, TI, and SA tasks, accuracy and macro-F1 metrics are used to evaluate predictive performance. For quantitative evaluation od RD task, we used a token-based, edit distance-based and sequence-based measure in the form of Jaccard Similarity (JS), Hamming distance (HD), and Ratcliff-Obershelp Similarity (ROS) metrics, respectively, as mentioned in \cite{ghosh2022cares}.

\subsection{Findings from Experiments}
Table~\ref{tab:result1} and ~\ref{tab:result2} illustrate the results of our proposed model, mExCB, and the other two variants of mExCB for Bully, Target, and Rationales detection tasks, respectively. Other baseline results are shown in Table~\ref{tab:result3}. From the tables containing results  we can conclude the following:

\textbf{(RQ1)} The proposed {\em mExCB} model outperforms all the baselines significantly for the CD task, improving 2.88\% accuracy over the best baseline, BERT-finetune.  For the rationale detection task, the improvement is 4.24\% in terms of ROS. This improvement in performance reveals the importance of utilizing some auxiliary tasks, sentiment analysis, rationale detection, and target identification in boosting the performance of the main task (CD). 

\textbf{(RQ2)} Unlike the four task variants (all tasks), Bully+Rationales+Sentiment (three tasks) settings attain the best results for CD and RD tasks. This finding established that increasing the number of tasks in a multitask framework does not always improve the performance of the main task compared to some less number task combinations. This decrease in the performance of the four tasks variant could be due to the task TI, which does not perform well and has a negative effect on other tasks.
\begin{table*}[hbt]
\centering
\caption{Experimental results of different multitask variants with BERT and VecMap embeddings for Bully and Target tasks.  SA: Sentiment Analysis, CD: Cyberbully Detection, RD: Rationales Detection, TI: Target Identification}
\label{tab:result1}
\scalebox{0.68}{
\begin{tabular}{l|l|cc|cc|cccc|cccc}
\hline
\multicolumn{1}{c|}{\multirow{3}{*}{\textbf{Embedding}}} &
  \multicolumn{1}{c|}{\multirow{3}{*}{\textbf{Model}}} &
  \multicolumn{2}{c|}{\textbf{Bully+Rationales}} &
  \multicolumn{2}{c|}{\textbf{\begin{tabular}[c]{@{}c@{}}Bully + Rationales\\ +Sentiment\end{tabular}}} &
  \multicolumn{4}{c|}{\textbf{Bully + Rationales+Target}} &
  \multicolumn{4}{c}{\textbf{Bully + Rationales+Sentiment+Target}} \\
\multicolumn{1}{c|}{} &
  \multicolumn{1}{c|}{} &
  \multicolumn{2}{c|}{\textbf{Bully}} &
  \multicolumn{2}{c|}{\textbf{Bully}} &
  \multicolumn{2}{c|}{\textbf{Bully}} &
  \multicolumn{2}{c|}{\textbf{Target}} &
  \multicolumn{2}{c|}{\textbf{Bully}} &
  \multicolumn{2}{c}{\textbf{Target}} \\
\multicolumn{1}{c|}{} &
  \multicolumn{1}{c|}{} &
  \textbf{Acc} &
  \textbf{macro F1} &
  \textbf{Acc} &
  \textbf{macro F1} &
  \textbf{Acc} &
  \multicolumn{1}{c|}{\textbf{macro F1}} &
  \textbf{Acc} &
  \textbf{macro F1} &
  \textbf{Acc} &
  \multicolumn{1}{c|}{\textbf{macro F1}} &
  \textbf{Acc} &
  \textbf{macro F1} \\ \hline
\multirow{3}{*}{VecMap} &
  mExCB$_{CNN}$ &
  79.93 &
  79.54 &
  80.76 &
  80.74 &
  80.43 &
  \multicolumn{1}{c|}{80.23} &
  50.86 &
  45.16 &
  80.37 &
  \multicolumn{1}{c|}{80.31} &
  51.97 &
  46.13 \\
 &
  mExCB$_{GRU}$ &
  79.52 &
  79.46 &
  80.43 &
  80.46 &
  80.26 &
  \multicolumn{1}{c|}{80.28} &
  51.12 &
  45.67 &
  80.26 &
  \multicolumn{1}{c|}{80.25} &
  51.40 &
  45.83 \\
 &
  mExCB &
  80.76 &
  80.66 &
  81.17 &
  81.15 &
  80.59 &
  \multicolumn{1}{c|}{80.36} &
  52.22 &
  44.37 &
  80.67 &
  \multicolumn{1}{c|}{80.53} &
  52.55 &
  49.05 \\ \hline
\multirow{3}{*}{mBERT} &
  mExCB$_{CNN}$ &
  80.87 &
  80.82 &
  81.51 &
  81.45 &
  81.09 &
  \multicolumn{1}{c|}{80.91} &
  50.93 &
  44.38 &
  80.84 &
  \multicolumn{1}{c|}{80.58} &
  52.31 &
  47.87 \\
 &
  mExCB$_{GRU}$ &
  80.67 &
  80.64 &
  81.25 &
  81.26 &
  81.99 &
  \multicolumn{1}{c|}{81.89} &
  51.47 &
  46.22 &
  79.93 &
  \multicolumn{1}{c|}{79.93} &
  51.81 &
  45.68 \\
 &
  mExCB &
  82.24 &
  82.31 &
  \textbf{83.31} &
  {\bf 83.24} &
  82.07 &
  \multicolumn{1}{c|}{82.11} &
  53.11 &
  49.08 &
  82.24 &
  \multicolumn{1}{c|}{82.19} &
  \textbf{54.54} &
  {\bf 50.20} \\ \hline
\end{tabular}}
\end{table*}

% \usepackage{multirow}
\begin{table}[hbt]
\centering
\caption{Experimental results of different multitask variants with BERT and VecMap embeddings for RD task. RD: Rationales Detection}
\label{tab:result2}
\scalebox{0.88}{
\begin{tabular}{l|l|crr|ccc|ccc|ccc}
\hline
\multicolumn{1}{c|}{\multirow{3}{*}{\textbf{Embedding}}} &
  \multicolumn{1}{c|}{\multirow{3}{*}{\textbf{Model}}} &
  \multicolumn{3}{c|}{\textbf{Bully+Rationales}} &
  \multicolumn{3}{c|}{\textbf{\begin{tabular}[c]{@{}c@{}}Bully + Rationales\\ +Sentiment\end{tabular}}} &
  \multicolumn{3}{c|}{\textbf{\begin{tabular}[c]{@{}c@{}}Bully + Rationales\\ +Target\end{tabular}}} &
  \multicolumn{3}{c}{\textbf{\begin{tabular}[c]{@{}c@{}}Bully + Rationale\\ +Target+Sentiment\end{tabular}}} \\ \cline{3-14} 
\multicolumn{1}{c|}{} &
  \multicolumn{1}{c|}{} &
  \multicolumn{3}{c|}{\textbf{Rationales}} &
  \multicolumn{3}{c|}{\textbf{Rationales}} &
  \multicolumn{3}{c|}{\textbf{Rationales}} &
  \multicolumn{3}{c}{\textbf{Rationales}} \\
\multicolumn{1}{c|}{} &
  \multicolumn{1}{c|}{} &
  \textbf{JS} &
  \multicolumn{1}{c}{\textbf{HD}} &
  \multicolumn{1}{l|}{\textbf{ROS}} &
  \textbf{JS} &
  \textbf{HD} &
  \textbf{ROS} &
  \textbf{JS} &
  \textbf{HD} &
  \textbf{ROS} &
  \textbf{JS} &
  \textbf{HD} &
  \textbf{ROS} \\ \hline
\multirow{3}{*}{VecMap} &
  mExCB$_{CNN}$ &
  44.11 &
  43.22 &
  44.18 &
  44.58 &
  43.41 &
  44.62 &
  44.31 &
  43.15 &
  45.34 &
  44.68 &
  43.57 &
  45.17 \\
 &
  mExCB$_{GRU}$ &
  44.74 &
  43.45 &
  45.86 &
  44.93 &
  43.69 &
  46.87 &
  44.41 &
  43.98 &
  46.23 &
  44.87 &
  43.69 &
  46.34 \\
 &
  mExCB &
  45.68 &
  43.98 &
  47.86 &
  45.71 &
  44.15 &
  48.64 &
  45.11 &
  44.19 &
  46.66 &
  44.73 &
  43.89 &
  46.47 \\ \hline
\multirow{3}{*}{mBERT} &
  mExCB$_{CNN}$ &
  45.34 &
  43.28 &
  47.58 &
  45.28 &
  43.78 &
  47.62 &
  45.44 &
  43.72 &
  47.22 &
  45.31 &
  43.56 &
  47.23 \\
 &
  mExCB$_{GRU}$ &
  45.37 &
  43.57 &
  47.89 &
  45.62 &
  43.96 &
  48.23 &
  45.53 &
  43.71 &
  47.75 &
  45.49 &
  43.88 &
  47.71 \\
 &
  mExCB &
  45.83 &
  44.26 &
  48.78 &
  \textbf{46.39} &
  \textbf{44.65} &
  \textbf{49.42} &
  45.74 &
  43.87 &
  47.67 &
  45.74 &
  43.67 &
  47.92 \\ \hline
\end{tabular}}
\end{table}
\begin{table}[hbt]
\centering
\caption{Results of different baseline methods  evaluated on {\em BullyExplain} data }
\label{tab:result3}
\begin{tabular}{l|lc|lclll}
\hline
\multicolumn{1}{c|}{\multirow{3}{*}{\textbf{Model}}} &
  \multicolumn{2}{c|}{\multirow{2}{*}{\textbf{Bully}}} &
  \multicolumn{5}{c}{\textbf{Bully+Rationales}} \\
\multicolumn{1}{c|}{} &
  \multicolumn{2}{c|}{} &
  \multicolumn{2}{c|}{\textbf{Bully}} &
  \multicolumn{3}{c}{\textbf{Rationales}} \\
\multicolumn{1}{c|}{} &
  \textbf{Acc} &
  \textbf{macro F1} &
  \textbf{Acc} &
  \multicolumn{1}{l|}{\textbf{macro F1}} &
  \textbf{JS} &
  \textbf{HD} &
  \textbf{ROS} \\ \hline
CNN GRU &
  78.78 &
  78.37 &
  79.28 &
  \multicolumn{1}{l|}{79.14} &
  43.78 &
  42.81 &
  43.77 \\
BiRNN &
  78.62 &
  78.55 &
  79.44 &
  \multicolumn{1}{l|}{79.51} &
  42.11 &
  42.53 &
  43.12 \\
$BiRNN_{attn}$ &
  79.52 &
  79.45 &
  79.93 &
  \multicolumn{1}{l|}{80.01} &
  43.15 &
  43.02 &
  44.38 \\
$BERT_{fine}$ &
  80.18 &
  80.08 &
  \textbf{80.43} &
  \multicolumn{1}{l|}{\bf 80.35} &
  \textbf{44.53} &
  {\bf 43.72} &
  {\bf 45.18} \\ \hline
\end{tabular}
\end{table}
 For the Target Identification (TI) task, the maximum accuracy and F1 score of 54.54 and 50.20, respectively, are attained. This low accuracy in the TI task could be due to the imbalanced nature of the Target class.   

\textbf{(RQ3)} {\em mExCB} has consistently performed better than mExCB$_{CNN}$ and mExCB$_{GRU}$ in both embedding strategies (BERT / VecMap). Like, in (CD+RD+SA) multi-task setting, {\em mExCB} performs better than mExCB$_{CNN}$ and mExCB$_{GRU}$ with improvements in the F1 score of 1.79\% and 1.98\%, respectively for the CD task. From Table~\ref{tab:result1}, we can notice that mExCB$_{CNN}$ performs better than mExCB$_{GRU}$ for the CD task most of the time, and the reverse scenario occurs for the RD task. That is why for the RD task, only the self-attenuated sub-sentence level Bi-GRU features have been sent to the RD output layer. This finding supports the idea of using both GRU and CNN so that model can learn both long-range semantic features as well as local key phrase-based information. 

\textbf{(RQ4)} We have already experimented with fine-tuning BERT~\cite{mathew2020hatexplain} (Baseline-4) and achieved 80.18\% and 80.43\% (see Table~\ref{tab:result3}) accuracy values in single task (CD) and multi-task (CD+RD) settings, respectively, for the CD task. On the other hand, without fine-tuning the BERT followed by CNN and GRU, we have obtained an accuracy of 83.31\% for the CD task with (CD+RD+SA) multi-task setting. The BERT fine-tuning version of our proposed model achieved the highest accuracy of 81.59\% for the CD task, which underperforms the non-fine-tuning version of our model. One of the possible reasons why fine-tuning is not performing well could be the less number of samples in our dataset. 

\textbf{(RQ5)} In Table~\ref{tab:result1}, we can observe that all the models performed better when BERT was used for embedding generation instead of VecMap. This result again highlights the superiority of the BERT model in different types of NLP tasks. We have not included the VecMap embedded results in Table \ref{tab:result3} containing baselines  as it performs poorly. 

We have conducted a statistical t-test on the results of ten different runs of our proposed model and other baselines and obtained a p-value less than 0.05.

\subsection{Comparison with SOTA} 
Table~\ref{tab:sota} shows the results of the existing state-of-the-art approach for CD and SA tasks on the Hindi-English code-mixed dataset. We keep the original two tasks,  SA and CD, in our proposed model, {\em mExCB} (CD+SA) for a fair comparison between SOTA. Table~\ref{tab:sota} shows that mExCB (CD+SA) also outperforms the SOTA, illustrating that our proposed model can perform better without the introduced annotations (RD and TI). Furthermore, when we add a new task RD, the three task combination variant, {\em mExCB }(CD+SA+RD) outperforms both SOTA (MT-BERT+VecMap)  and {\em mExCB}(CD+SA) with  improvements in the accuracy value of 2.19\% and 1.36\%, respectively, for the CD task. This improvement illustrates the importance of incorporating the RD task in enhancing the performance of the main task, i.e., CD. Hence, our proposed approach is beneficial from both aspects, (i) Enhanced the performance of the CD task and (ii) Generates a human-like explanation to support the model's decision, which is vital in any AI-based task.

\begin{table}[hbt]
\centering
\caption{Results of  state-of-the-art model and the proposed model; ST: Single Task, MT: Multi-Task}
\label{tab:sota}
\begin{tabular}{lcccc}
\hline
\multicolumn{1}{l|}{}                       & \multicolumn{2}{c|}{\textbf{Bully}} & \multicolumn{2}{c}{\textbf{Sentiment}} \\
\multicolumn{1}{l|}{\multirow{-2}{*}{\textbf{Model}}}     & \textbf{Acc}   & \multicolumn{1}{c|}{\textbf{F1}}                  & \textbf{Acc}   & \textbf{F1}    \\ \hline
\multicolumn{5}{c}{\textbf{SOTA}}                                                                                          \\ \hline
\multicolumn{1}{l|}{BERT\_finetune~\cite{mathew2020hatexplain}}         & 80.18  & \multicolumn{1}{c|}{80.08} & 76.10              & 75.62             \\
\multicolumn{1}{l|}{ST- BERT+VecMap~\cite{maity2021multi}}        & 79.97  & \multicolumn{1}{c|}{80.13} & 75.53              & 75.38             \\
\multicolumn{1}{l|}{MT-BERT+VecMap~\cite{maity2021multi}}         & 81.12  & \multicolumn{1}{c|}{81.50} & 77.46              & 76.95             \\ \hline
\multicolumn{5}{c}{\textbf{Ours}}                                                                                          \\ \hline
\multicolumn{1}{l|}{\textbf{mExCB (CD+SA)}} & 81.95  & \multicolumn{1}{c|}{82.04} & 77.55              & 77.12             \\
\multicolumn{1}{l|}{\textbf{mExCB (CD+SA+RD)}}            & \textbf{83.31} & \multicolumn{1}{c|}{\textbf{83.24}}               & \textbf{78.54} & \textbf{78.13} \\ \hline
\rowcolor[HTML]{ECF4FF} 
\multicolumn{1}{l|}{\cellcolor[HTML]{ECF4FF}Improvements} & 2.19           & \multicolumn{1}{c|}{\cellcolor[HTML]{ECF4FF}1.74} & 1.08           & 1.18           \\ \hline
\end{tabular}
\end{table}

\begin{table}[hbt]
\caption{In comparison to human annotators, rationales identified by several models are shown. Green highlights indicate  agreements between the human annotator and the model. Orange highlighted tokens are predicted by models, not by human annotators. Yellow highlighted tokens are predicted by models but are not present in the original text. }
\label{tab:error}
\scalebox{0.75}{
\begin{tabular}{l|l|l}
\hline
\textbf{Model}                 & \textbf{Text}                                                                              & \textbf{Bully Label} \\ \hline
Human annotator (T1)      & Semi final tak usi bnde ne pahochaya hai jisko tu gandu bol raha . & Non-Bully        \\ 
\textbf{Translation}           &\textbf{The person you are calling ass*ole is the one that helped us to reach the semi finals.}  &      \\ 
CNN-GRU              &   \colorbox{orange}{Semi final} tak usi bnde ne pahochaya hai jisko tu \colorbox{orange}{gandu} bol raha . & Bully    \\ 
BiRNN-Attn      & Semi final tak usi bnde ne pahochaya hai jisko \colorbox{orange}{tu gandu} bol raha .   & Bully   \\ 
mExCB$_{CNN}$     & Semi final tak usi bnde ne pahochaya hai jisko tu \colorbox{orange}{gandu} bol raha . & Bully        \\ 
mExCB$_{GRU}$ & Semi final tak usi bnde ne pahochaya hai jisko tu \colorbox{orange}{gandu} bol raha . & Bully \\ 
mExCB        & Semi final tak usi bnde ne pahochaya hai jisko tu \colorbox{orange}{gandu} bol raha . & Bully        \\ 
Human annotator (T2)      & Abey \colorbox{green}{mc} gb road r ki pehle customer ki \colorbox{green}{najayaz auladte}                            & Bully        \\ 
\textbf{Translation}           & \textbf{You are the illegitimate children of first customer of GB road.}  &          \\ 
CNN-GRU               & Abey mc \colorbox{orange}{gb road r ki} pehle customer ki najayaz auladte  &   Bully       \\ 
BiRNN -Attn      & Abey mc gb \colorbox{orange}{road} r \colorbox{orange}{ki} pehle customer ki \colorbox{green}{najayaz} auladte   &    Bully      \\ 
mExCB$_{CNN}$ & Abey \colorbox{green}{mc} gb road r ki \colorbox{orange}{pehle customer} ki \colorbox{green}{najayaz} auladte                          & Bully        \\ \

mExCB$_{GRU}$        &\colorbox{orange}{Abey} \colorbox{green}{mc} gb road r ki pehle \colorbox{orange}{customer} ki \colorbox{green}{najayaz auladte}                            & Bully        \\ 
mExCB    & \colorbox{orange}{Abey} \colorbox{green}{mc} gb road r ki pehle customer ki \colorbox{green}{najayaz} \colorbox{green}{auladte}                  & Bully        \\ \hline
\end{tabular}
}
\end{table}

\section{Error Analysis}
We have manually checked some samples from the test set to examine how machine-generated rationales and bully labels differ from the human annotator's decision. Table~\ref{tab:error} shows the predicted rationales and bully labels of a few test samples obtained by different baselines (CNN-GRU, BiRNN-Attn) and our proposed models (mExCB). 
\begin{enumerate}
    \item It can be observed that the human annotator labeled the T1 tweet as Non-Bully. In contrast, all the models (both baselines and $GenEx$ models) predicted the label as Bully highlighting the offensive word \textit{gandu} (Asshole), supporting their predictions. This shows that the model cannot comprehend the context of this offensive word as it is not directed at anyone and has been used more in a sarcastic manner, highlighting the model's limitation in understanding indirect and sarcastic statements. 

    \item  All the models (both baselines and our proposed models) predicted the correct label for tweet T2. But if we see the rationales predicted (highlighted part), none of the baseline models performed well compared to human decisions. Both {\em mExCB$_{GRU}$} and {\em mExCB} can predict all the words present in the ground truth rationale, but it also predicts other phrases as the rationale. In this case, we can also notice that {\em mExCB$_{GRU}$} predicts some more tokens which are not rationales compared to {\em mExCB} model. Our proposed model performs slightly better than others in the RD task. Models that excel in classification may not always be able to give reasonable and accurate rationales for their decisions as classification task needs sentence-level features~\cite{mathew2020hatexplain}. In contrast, rationales detection mainly focuses on the token-level feature. This observation (low performance in RD task) indicates that more research is needed on explainability. We believe our developed dataset will help future research on explainable cyberbullying detection.
\end{enumerate}

\section{Conclusion and Future Work}
In this paper we have made an attempt in solving the cyberbullying detection task in code-mixed setting keeping the explainability aspect in mind. As explainable AI systems help in improving  trustworthiness and confidence while deployed in real-time and cyberbully detection systems are required to be installed in different social media sites for online monitoring, generating rationales behind taken decisions is a must. The contributions of the current work are two fold: (a) developing the first explainable cyberbully detection dataset in code-mixed language where rationales/phrases used for decision making are annotated along with bully label, sentiment label and target label. (b) A multitask framework {\em mExCB} based on word and sub-sentence label attention has been proposed to solve four tasks (Bully, Sentiment, Rationales, Target) simultaneously. Our proposed model outperforms all the baselines and beats state-of-the-art with an accuracy score of 2.19\% for cyberbullying detection. 

In future we would like to work on explainable cyberbully detection from code-mixed data considering image and text modalities.

{\bf Acknowledgement} Dr. Sriparna Saha gratefully acknowledges the Young Faculty Research Fellowship (YFRF) Award, supported by Visvesvaraya Ph.D. Scheme for Electronics and IT, Ministry of Electronics and Information Technology (MeitY), Government of India, being implemented by Digital India Corporation (formerly Media Lab Asia) for carrying
out this research. The Authors would also like to acknowledge the support of Ministry of Home Affairs (MHA), India, for conducting
this research.

\bibliographystyle{splncs04}
\bibliography{custom}

\begin{thebibliography}{10}
\providecommand{\url}[1]{\texttt{#1}}
\providecommand{\urlprefix}{URL }
\providecommand{\doi}[1]{https://doi.org/#1}

\bibitem{agrawal2018deep}
Agrawal, S., Awekar, A.: Deep learning for detecting cyberbullying across
  multiple social media platforms. In: European conference on information
  retrieval. pp. 141--153. Springer (2018)

\bibitem{artetxe2017learning}
Artetxe, M., Labaka, G., Agirre, E.: Learning bilingual word embeddings with
  (almost) no bilingual data. In: Proceedings of the 55th Annual Meeting of the
  Association for Computational Linguistics (Volume 1: Long Papers). pp.
  451--462 (2017)

\bibitem{rwml4}
Badjatiya, P., Gupta, S., Gupta, M., Varma, V.: Deep learning for hate speech
  detection in tweets. In: Proceedings of the 26th International Conference on
  World Wide Web Companion. pp. 759--760 (2017)

\bibitem{rwml5}
Balakrishnan, V., Khan, S., Arabnia, H.R.: Improving cyberbullying detection
  using twitter users’ psychological features and machine learning. Computers
  \& Security  \textbf{90},  101710 (2020)

\bibitem{rwcm3}
Bohra, A., Vijay, D., Singh, V., Akhtar, S.S., Shrivastava, M.: A dataset of
  hindi-english code-mixed social media text for hate speech detection. In:
  Proceedings of the second workshop on computational modeling of people’s
  opinions, personality, and emotions in social media. pp. 36--41 (2018)

\bibitem{camburu2018snli}
Camburu, O.M., Rockt{\"a}schel, T., Lukasiewicz, T., Blunsom, P.: e-snli:
  Natural language inference with natural language explanations. Advances in
  Neural Information Processing Systems  \textbf{31} (2018)

\bibitem{cho2014properties}
Cho, K., Van~Merri{\"e}nboer, B., Bahdanau, D., Bengio, Y.: On the properties
  of neural machine translation: Encoder-decoder approaches. arXiv preprint
  arXiv:1409.1259  (2014)

\bibitem{dadvar2014experts}
Dadvar, M., Trieschnigg, D., Jong, F.d.: Experts and machines against bullies:
  A hybrid approach to detect cyberbullies. In: Canadian conference on
  artificial intelligence. pp. 275--281. Springer (2014)

\bibitem{devlin2018bert}
Devlin, J., Chang, M.W., Lee, K., Toutanova, K.: Bert: Pre-training of deep
  bidirectional transformers for language understanding. arXiv preprint
  arXiv:1810.04805  (2018)

\bibitem{deyoung2019eraser}
DeYoung, J., Jain, S., Rajani, N.F., Lehman, E., Xiong, C., Socher, R.,
  Wallace, B.C.: Eraser: A benchmark to evaluate rationalized nlp models. arXiv
  preprint arXiv:1911.03429  (2019)

\bibitem{rwml1}
Dinakar, K., Reichart, R., Lieberman, H.: Modeling the detection of textual
  cyberbullying. In: Proceedings of the International Conference on Weblog and
  Social Media 2011. Citeseer (2011)

\bibitem{fleiss1971measuring}
Fleiss, J.L.: Measuring nominal scale agreement among many raters.
  Psychological bulletin  \textbf{76}(5), ~378 (1971)

\bibitem{ghosh2022cares}
Ghosh, S., Roy, S., Ekbal, A., Bhattacharyya, P.: Cares: Cause recognition for
  emotion in suicide notes. In: European Conference on Information Retrieval.
  pp. 128--136. Springer (2022)

\bibitem{grave2018learning}
Grave, E., Bojanowski, P., Gupta, P., Joulin, A., Mikolov, T.: Learning word
  vectors for 157 languages. arXiv preprint arXiv:1802.06893  (2018)

\bibitem{gunning2019xai}
Gunning, D., Stefik, M., Choi, J., Miller, T., Stumpf, S., Yang, G.Z.:
  Xai—explainable artificial intelligence. Science Robotics  \textbf{4}(37),
  eaay7120 (2019)

\bibitem{rwcm5}
Kamble, S., Joshi, A.: Hate speech detection from code-mixed hindi-english
  tweets using deep learning models. arXiv preprint arXiv:1811.05145  (2018)

\bibitem{karim2021deephateexplainer}
Karim, M.R., Dey, S.K., Islam, T., Sarker, S., Menon, M.H., Hossain, K.,
  Hossain, M.A., Decker, S.: Deephateexplainer: Explainable hate speech
  detection in under-resourced bengali language. In: 2021 IEEE 8th
  International Conference on Data Science and Advanced Analytics (DSAA). pp.
  1--10. IEEE (2021)

\bibitem{kim2014convolutional}
Kim, Y.: Convolutional neural networks for sentence classification. arXiv
  preprint arXiv:1408.5882  (2014)

\bibitem{kumar2018aggression}
Kumar, R., Reganti, A.N., Bhatia, A., Maheshwari, T.: Aggression-annotated
  corpus of hindi-english code-mixed data. arXiv preprint arXiv:1803.09402
  (2018)

\bibitem{rwcm2}
Kumar, R., Reganti, A.N., Bhatia, A., Maheshwari, T.: Aggression-annotated
  corpus of hindi-english code-mixed data. arXiv preprint arXiv:1803.09402
  (2018)

\bibitem{lewis2010handbook}
Lewis, M., Haviland-Jones, J.M., Barrett, L.F.: Handbook of emotions. Guilford
  Press (2010)

\bibitem{maity2021bert}
Maity, K., Saha, S.: Bert-capsule model for cyberbullying detection in
  code-mixed indian languages. In: International Conference on Applications of
  Natural Language to Information Systems. pp. 147--155. Springer (2021)

\bibitem{maity2021multi}
Maity, K., Saha, S.: A multi-task model for sentiment aided cyberbullying
  detection in code-mixed indian languages. In: International Conference on
  Neural Information Processing. pp. 440--451. Springer (2021)

\bibitem{maity2022mtbullygnn}
Maity, K., Sen, T., Saha, S., Bhattacharyya, P.: Mtbullygnn: A graph neural
  network-based multitask framework for cyberbullying detection. IEEE
  Transactions on Computational Social Systems  (2022)

\bibitem{mathew2020hatexplain}
Mathew, B., Saha, P., Yimam, S.M., Biemann, C., Goyal, P., Mukherjee, A.:
  Hatexplain: A benchmark dataset for explainable hate speech detection. arXiv
  preprint arXiv:2012.10289  (2020)

\bibitem{cm1}
Myers-Scotton, C.: Duelling languages: Grammatical structure in codeswitching.
  Oxford University Press (1997)

\bibitem{ncrb}
NCRB: Crime in india – 2020. National Crime Records Bureau  (2020)

\bibitem{paul2020cyberbert}
Paul, S., Saha, S.: Cyberbert: Bert for cyberbullying identification.
  Multimedia Systems pp.~1--8 (2020)

\bibitem{pramanick2021detecting}
Pramanick, S., Dimitrov, D., Mukherjee, R., Sharma, S., Akhtar, M., Nakov, P.,
  Chakraborty, T., et~al.: Detecting harmful memes and their targets. arXiv
  preprint arXiv:2110.00413  (2021)

\bibitem{rajani2019explain}
Rajani, N.F., McCann, B., Xiong, C., Socher, R.: Explain yourself! leveraging
  language models for commonsense reasoning. arXiv preprint arXiv:1906.02361
  (2019)

\bibitem{regulation2016regulation}
Regulation, P.: Regulation (eu) 2016/679 of the european parliament and of the
  council. Regulation (eu)  \textbf{679}, ~2016 (2016)

\bibitem{rwml2}
Reynolds, K., Kontostathis, A., Edwards, L.: Using machine learning to detect
  cyberbullying. In: 2011 10th International Conference on Machine learning and
  applications and workshops. vol.~2, pp. 241--244. IEEE (2011)

\bibitem{rijhwani2017estimating}
Rijhwani, S., Sequiera, R., Choudhury, M., Bali, K., Maddila, C.S.: Estimating
  code-switching on twitter with a novel generalized word-level language
  detection technique. In: Proceedings of the 55th annual meeting of the
  association for computational linguistics (volume 1: long papers). pp.
  1971--1982 (2017)

\bibitem{saha2021multitask}
Saha, T., Upadhyaya, A., Saha, S., Bhattacharyya, P.: A multitask multimodal
  ensemble model for sentiment-and emotion-aided tweet act classification. IEEE
  Transactions on Computational Social Systems  (2021)

\bibitem{singh2021multitask}
Singh, A., Saha, S., Hasanuzzaman, M., Dey, K.: Multitask learning for
  complaint identification and sentiment analysis. Cognitive Computation pp.
  1--16 (2021)

\bibitem{cyb1}
Smith, P.K., Mahdavi, J., Carvalho, M., Fisher, S., Russell, S., Tippett, N.:
  Cyberbullying: Its nature and impact in secondary school pupils. Journal of
  child psychology and psychiatry  \textbf{49}(4),  376--385 (2008)

\bibitem{sticca2013longitudinal}
Sticca, F., Ruggieri, S., Alsaker, F., Perren, S.: Longitudinal risk factors
  for cyberbullying in adolescence. Journal of community \& applied social
  psychology  \textbf{23}(1),  52--67 (2013)

\bibitem{vaswani2017attention}
Vaswani, A., Shazeer, N., Parmar, N., Uszkoreit, J., Jones, L., Gomez, A.N.,
  Kaiser, {\L}., Polosukhin, I.: Attention is all you need. In: Advances in
  neural information processing systems. pp. 5998--6008 (2017)

\bibitem{waseem2016hateful}
Waseem, Z., Hovy, D.: Hateful symbols or hateful people? predictive features
  for hate speech detection on twitter. In: Proceedings of the NAACL student
  research workshop. pp. 88--93 (2016)

\bibitem{ybarra2006examining}
Ybarra, M.L., Mitchell, K.J., Wolak, J., Finkelhor, D.: Examining
  characteristics and associated distress related to internet harassment:
  findings from the second youth internet safety survey. Pediatrics
  \textbf{118}(4),  e1169--e1177 (2006)

\bibitem{yin2017comparative}
Yin, W., Kann, K., Yu, M., Sch{\"u}tze, H.: Comparative study of cnn and rnn
  for natural language processing. arXiv preprint arXiv:1702.01923  (2017)

\bibitem{zaidan2007using}
Zaidan, O., Eisner, J., Piatko, C.: Using “annotator rationales” to improve
  machine learning for text categorization. In: Human language technologies
  2007: The conference of the North American chapter of the association for
  computational linguistics; proceedings of the main conference. pp. 260--267
  (2007)

\bibitem{zhang2018generalized}
Zhang, Z., Sabuncu, M.: Generalized cross entropy loss for training deep neural
  networks with noisy labels. Advances in neural information processing systems
   \textbf{31} (2018)

\end{thebibliography}
\end{document}